\IfFileExists{ieeeconf.cls}{%
  \documentclass[letterpaper, 10 pt, conference]{ieeeconf}%
  \IEEEoverridecommandlockouts
  \overrideIEEEmargins
}{%
  \documentclass[letterpaper,10pt,journal,twoside]{IEEEtran}%
}
\makeatletter
\usepackage{iftex}
\ifXeTeX\usepackage{fontspec}\fi

\providecommand{\bstctlcite}[1]{\@bsphack\if@filesw\immediate\write\@auxout{\string\citation{#1}}\fi\@esphack}
\@ifundefined{IEEEkeywords}{%
  \newenvironment{IEEEkeywords}{\par\noindent\textbf{\textit{Index Terms}---}}{\par}}{}
\makeatother

\usepackage{cite}
\usepackage{amsmath,amssymb}
\usepackage{graphicx}
\usepackage{booktabs}
\usepackage{multirow}
\usepackage{array}
\usepackage[table]{xcolor}
\usepackage{xspace}
\usepackage{url}
\usepackage[hidelinks]{hyperref}
\usepackage{letltxmacro}
\newif\ifnamed\namedfalse
\namedtrue
\newcommand{\authornames}{Yang Liu, Xiaoli Liu, Liuzhao Li, Yuduo Song, Muyuan Liu, Xiang Gao}
\ifnamed
\hypersetup{pdfauthor={\authornames}}
\else
\hypersetup{pdfauthor={Anonymous Authors}}
\fi

\usepackage[protrusion=true,expansion=false]{microtype}
\makeatletter
\let\gawm@nobreakdash\nobreakdash
\DeclareRobustCommand{\nobreakdash}{\gawm@nobreakdash}
\AtBeginDocument{\LetLtxMacro\gawm@ref\ref\DeclareRobustCommand{\ref}[1]{\mbox{\gawm@ref{#1}}}}
\makeatother

\definecolor{tbdorange}{RGB}{221,110,0}
\definecolor{oursbg}{RGB}{226,240,247}
\definecolor{groupbg}{RGB}{242,242,242}

\newif\ifdraftmarks\draftmarkstrue
\draftmarksfalse

\newcommand{\method}{GAWM\xspace}

\DeclareRobustCommand{\todo}[2][]{\ifdraftmarks\textcolor{red}{\textbf{[%
  \if\relax\detokenize{#1}\relax TODO\else #1\fi
  \if\relax\detokenize{#2}\relax\else: #2\fi]}}\else #2\fi}

\DeclareRobustCommand{\note}[2][]{\ifdraftmarks\textcolor{red}{\textbf{[%
  \if\relax\detokenize{#1}\relax NOTE\else #1\fi: #2]}}\else\expandafter\ignorespaces\fi}

\newcommand{\ours}{\rowcolor{oursbg}}

\definecolor{markgreen}{RGB}{47,125,79}
\definecolor{markred}{RGB}{192,57,75}
\newcommand{\cmark}{\textcolor{markgreen}{$\checkmark$}}
\newcommand{\xmark}{\textcolor{markred}{$\times$}}
\newcommand{\best}[1]{\textbf{#1}}
\newcommand{\second}[1]{\underline{#1}}
\newcommand{\tsup}[1]{\rlap{\smash{$^{#1}$}}}
\newcommand{\dz}{\phantom{0}}
\newcommand{\grouprow}[2]{\rowcolor{groupbg}\multicolumn{#1}{l}{\textsc{#2}}\\}

\newcommand{\ns}{n/s}

\begin{document}
\bstctlcite{IEEEexample:BSTcontrol}

\title{\method: A Compact Latent World-Action Model\\ for Multi-Task Robot Manipulation}

\ifnamed
\author{\authornames%
\thanks{All authors are with Genisom AI, Beijing, China. Corresponding author: Xiang Gao
(e-mail: gao.xiang.thu@gmail.com).}}
\else
\author{Anonymous Authors}
\fi

\maketitle

\begin{abstract}
World-action models (WAMs) incorporate the prediction of future observations into robot policy learning. However, most WAMs build on billion-parameter vision-language or video-generation backbones, which incur high training cost and inference latency. In this letter, we present \method{} (Geometry-Aware World Model), a compact multi-task WAM that predicts future observations in the latent space of a frozen DINOv3 encoder. \method{} compresses the features of each camera view into a small set of tokens. A lightweight predictor, conditioned on the task rather than on actions, forecasts these tokens at two future horizons. The forecast is supervised with the dense patch features of future frames, which requires it to encode the spatial layout of the future scene. An action decoder directly regresses a chunk of future actions from the current and predicted tokens. With fewer than 100M parameters and no robot or video pretraining, \method{} outperforms billion-parameter WAMs on RoboTwin~2.0 and performs comparably to billion-parameter VLAs on LIBERO. On a dual-arm robot, it achieves lower action-prediction error on held-out demonstrations than a fine-tuned billion-parameter VLA, with substantially lower inference latency. Ablation studies show that the latent world model substantially improves performance when training data or training time is limited. \ifnamed Code is available at \url{https://github.com/gaoxiang12/ga-wm}.\else Code will be released upon acceptance.\fi

\end{abstract}

\begin{IEEEkeywords}
Compact Models, Imitation Learning, Joint-Embedding Predictive Architectures, Latent World Models, Multi-Task Manipulation, Reactive Policies, World-Action Models
\end{IEEEkeywords}

\section{Introduction}\label{sec:intro}

Vision-language-action models (VLAs) have become a dominant paradigm for robot manipulation, transferring web-scale vision-language pretraining to action generation~\cite{brohan2023rt2,kim2024openvla,black2025pi0,intelligence2025pi05}. Most VLAs, however, map the current observation directly to actions, without explicit supervision on how the scene will evolve. World-action models (WAMs) address this limitation by additionally predicting future observations~\cite{du2023unipi,wu2023gr1,tian2024seer,li2026lingbotva,yuan2026fastwam,chen2026lawam,yang2026lilawam}. Some WAMs employ the prediction only as an auxiliary training target, whereas others condition the action generator on it at test time. Such foresight can provide the policy with an explicit representation of task progress, which is most beneficial when test scenes differ from the training data, e.g., under visual perturbations~\cite{gu2026arps,zhao2026fasterwam,liu2026plawvla}.

\begin{figure*}[t]
\centering
\includegraphics[width=\textwidth]{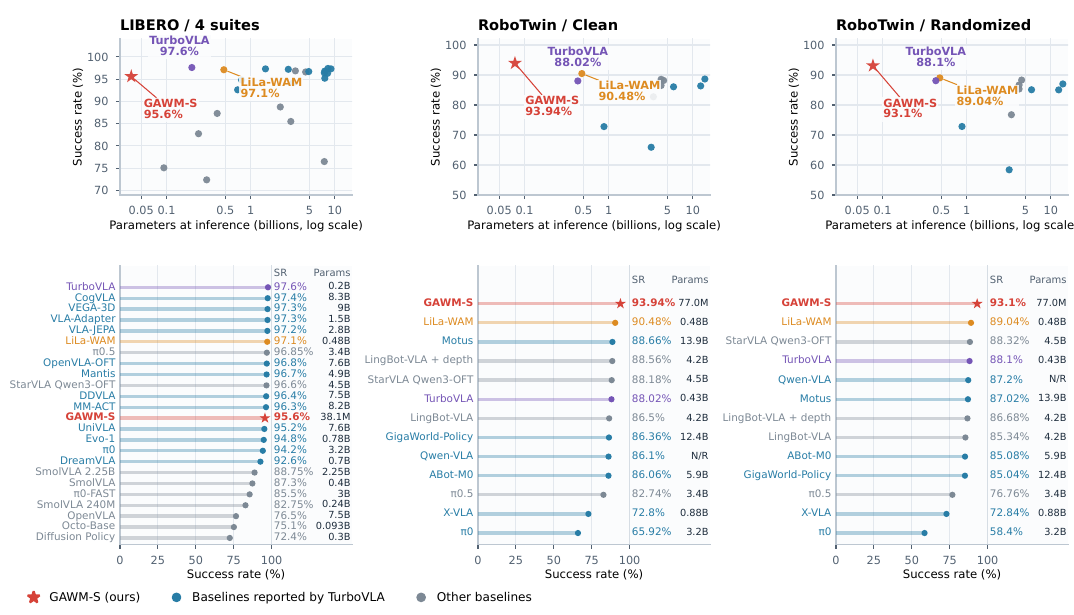}
\caption{\textbf{Success rate versus model size} on LIBERO (left; \method{}\nobreakdash-S with additional in-domain data) and on the clean (middle) and randomized (right) settings of RoboTwin~2.0. Top: success rate versus size. Bottom: methods ranked by success rate (SR) with size (as in Table~\ref{tab:main}, or as reported; N/R: not reported). Baselines (references in Table~\ref{tab:main} and~\cite{chi2023diffusion,pertsch2025fast,bu2025univla,li2025cogvla,yang2025mantis,liang2025mmact,sun2026vlajepa,wu2026vega3d,liang2025ddvla,zhang2025dreamvla,wu2026lingbotvla,ye2026gigaworld,yang2026abotm0,wang2026qwenvla}): blue, as reported by TurboVLA~\cite{xie2026turbovla}; gray, other sources. On RoboTwin~2.0, \method{}\nobreakdash-S is the smallest and best-performing of the plotted methods.}
\label{fig:teaser}
\end{figure*}

Most existing WAMs derive this foresight from large backbones. LingBot\nobreakdash-VA, Motus and Fast\nobreakdash-WAM build on the 5B-parameter video generator Wan~2.2~\cite{wan2025wan,li2026lingbotva,bi2025motus,yuan2026fastwam}, and LaWAM on a 2.1B-parameter VLA backbone~\cite{chen2026lawam}. The WAMs built on video generators predict future video frames and thus devote much of their capacity to appearance details irrelevant to control. Generating these frames also requires iterative denoising, which takes 4.5~s per query for LingBot\nobreakdash-VA~\cite{chen2026lawam}. Such models are costly to train and difficult to deploy on onboard computers.

Many deployed robots lack the onboard computational resources to run such models. They typically perform a fixed set of tasks repeatedly, for which reliable, low-latency execution matters more than understanding open-ended instructions. Compact policies are well suited to this setting. They operate at high control rates on onboard computers and can be trained at low cost for a new task set. On closed task sets, their success rates are comparable to those of billion-parameter VLAs~\cite{xie2026turbovla}. This motivates a hierarchical design~\cite{ahn2022saycan,shi2025hirobot}, in which a language model interprets instructions and selects the task, and a compact policy executes it. This letter focuses on the compact policy and asks whether it can benefit from foresight without a large backbone.

Joint-embedding predictive architectures (JEPAs) offer a lightweight route to foresight, as they predict representations instead of pixels~\cite{assran2023ijepa,bardes2024vjepa}. Compact JEPA world models such as LeWM~\cite{maes2026leworldmodel} and that of Liu et~al.~\cite{liu2026groundedjepa} rely on a small predictor. Their predictors, however, are conditioned on actions and support a planner that searches over candidate action sequences toward a goal image. Such planning not only limits the control rate, as the predictor must evaluate many candidates at every control step, but also requires a goal image to specify each task. A reactive policy, in contrast, avoids both costs by computing actions without search, which in turn requires a predictor that forecasts the future implied by the task rather than by candidate actions.

\begin{figure*}[t]
\centering
\includegraphics[width=\textwidth]{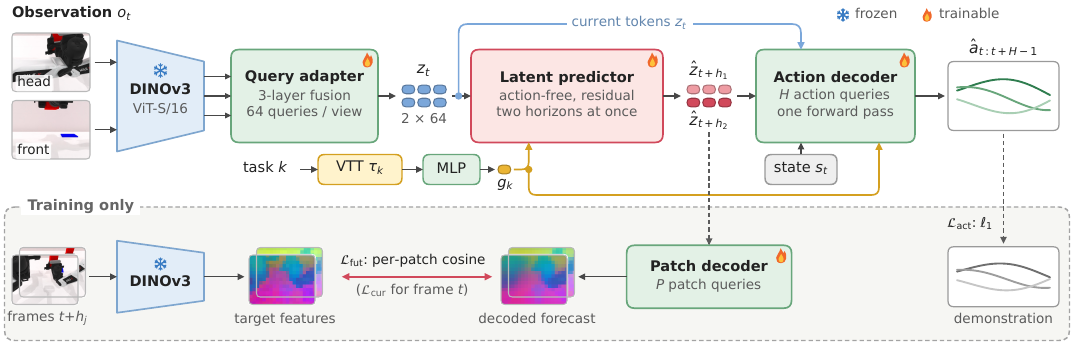}
\caption{\textbf{Overview of \method{}.} A frozen DINOv3 encoder and a query adapter encode each camera view into 64 visual tokens, stacked into $z_t$. The latent predictor, conditioned on the task code $g_k$, forecasts these tokens at two future horizons, and the action decoder directly regresses a chunk of future actions from the current tokens, the forecast, the robot state and the task code. Dashed panel: training only; a patch decoder maps the forecast to the encoder's patch features of the recorded future frames (thumbnails as in Fig.~\ref{fig:pca}).}
\label{fig:arch}
\end{figure*}

We present \method{} (Geometry-Aware World Model), a compact multi-task WAM that extends JEPA-style latent prediction to reactive control (Fig.~\ref{fig:arch}). \method{} encodes each camera view into 64 tokens from multi-layer features of a frozen DINOv3 ViT\nobreakdash-S encoder~\cite{simeoni2025dinov3}, as in LiLa\nobreakdash-WAM~\cite{yang2026lilawam}. A lightweight predictor, conditioned on a task code rather than on actions, forecasts these tokens at two future horizons in parallel. An action decoder in the style of Action Chunking with Transformers (ACT)~\cite{zhao2023act,xie2026turbovla} regresses an action chunk from the current tokens, the forecast, the robot state and the task code. As in LiLa\nobreakdash-WAM, the forecast is supervised with the frozen encoder's patch features of future frames. This dense target requires the forecast to encode the future positions of the gripper and objects. The target comes from a frozen encoder and therefore cannot collapse, removing the need for a regularizer such as SIGReg~\cite{balestriero2025lejepa}. \method{} employs neither a vision-language model (VLM) nor a video generator, which dominate the size of existing WAMs.

With 77M parameters and no robot or video pretraining, \method{}\nobreakdash-S, the ViT\nobreakdash-S variant, achieves success rates of 93.9\% and 93.1\% in the clean and randomized settings of RoboTwin~2.0~\cite{chen2025robotwin2} (Fig.~\ref{fig:teaser}), outperforming billion-parameter WAMs (Table~\ref{tab:main}). On LIBERO~\cite{liu2023libero}, it attains 95.6\% with 38M parameters, above the 3.2B-parameter $\pi_0$. On a dual-arm robot, \method{}\nobreakdash-S achieves lower action-prediction error on held-out demonstrations than a fine-tuned billion-parameter VLA, with substantially lower inference latency. These results show that, on a fixed task set, a policy can achieve high success without language input. Ablations show that the world model yields large gains when training data or training time is limited.

The main contributions of this letter are summarized as follows:
\begin{itemize}
\item We propose \method{}, a compact JEPA-style WAM for reactive control, in which a task-conditioned predictor forecasts the latent tokens of a frozen encoder and an action decoder regresses actions from this forecast.
\item We show that, without a VLM or a video generator, a multi-task policy with fewer than 100M parameters exceeds billion-parameter WAMs on RoboTwin~2.0, and predicts real-robot actions more accurately than a fine-tuned billion-parameter VLA.
\item Through ablations, we show that the world model yields large gains, especially when training data or training time is limited. Further analyses show that the forecast encodes the future spatial layout of the scene and that the action decoder attends to it in temporal order.
\end{itemize}

\section{Related Work}\label{sec:related}

\textbf{Vision-language-action models.} VLAs build on billion-parameter vision-language models pretrained on web-scale data and fine-tune them on robot data~\cite{brohan2023rt2,kim2024openvla,black2025pi0,intelligence2025pi05}. Subsequent work improves their action generation with parallel decoding~\cite{kim2025oft}, flow matching~\cite{black2025pi0} or action tokenization~\cite{pertsch2025fast}. These models map the current observation and the instruction directly to actions. On closed task sets, however, the language instruction contributes little to task success. Policies conditioned on the instruction largely ignore it under perturbations~\cite{zhou2025liberopro}. Similarly, replacing the instruction with a task identifier reduces the success of TurboVLA by only one point~\cite{xie2026turbovla}. LiLa\nobreakdash-WAM~\cite{yang2026lilawam} conditions its policy on a visual transition token computed from each task's demonstrations instead of the instruction. \method{} adopts it as its task code.

\textbf{World-action models.} WAMs in pixel space predict future frames, either as an auxiliary output of the policy~\cite{wu2023gr1} or as the input to an inverse-dynamics model~\cite{du2023unipi,tian2024seer}. More recent pixel-space WAMs build on pretrained video generators~\cite{li2026lingbotva,bi2025motus,yuan2026fastwam}. Latent WAMs instead predict learned representations. DreamVLA~\cite{zhang2025dreamvla} forecasts dynamic, depth and semantic features, and LaWAM~\cite{chen2026lawam} conditions a 2.1B-parameter policy on DINOv3 features predicted by a latent world model. LiLa\nobreakdash-WAM (0.48B) decodes foresight tokens from its action transformer and supervises them with the patch features of a frozen multi-layer DINOv3 front-end.

WAMs also differ in the role of the prediction at test time. Fast\nobreakdash-WAM~\cite{yuan2026fastwam} and LiLa\nobreakdash-WAM use it only as a training signal and generate no future at inference. Faster\nobreakdash-WAM~\cite{zhao2026fasterwam}, in contrast, shows that conditioning the policy on the predicted future at test time is critical under distribution shift. PLaW\nobreakdash-VLA~\cite{liu2026plawvla} and ARPS~\cite{gu2026arps} also condition the policy on a predicted latent future, but build on a $\pi_{0.5}$-scale policy and a video WAM, respectively. \method{} brings test-time future conditioning to a compact multi-task policy. It adopts the front-end and dense target of LiLa\nobreakdash-WAM but delegates future prediction to a dedicated task-conditioned predictor with 4\nobreakdash--6M parameters, and its action decoder is conditioned on the forecast at every step. The forecast describes the scenes at the midpoint and end of the action chunk, and the small predictor adds little inference cost.

\textbf{JEPA world models.} V\nobreakdash-JEPA~2\nobreakdash-AC~\cite{assran2025vjepa2} trains an action-conditioned predictor on frozen video features, and DINO\nobreakdash-WM~\cite{zhou2024dinowm} on frozen DINOv2 features. Both plan toward a goal image. LeWM~\cite{maes2026leworldmodel} trains a compact encoder and predictor end to end from pixels and prevents representation collapse with the SIGReg regularizer of LeJEPA~\cite{balestriero2025lejepa}. Liu et~al.~\cite{liu2026groundedjepa} instead regularize the same framework with inverse-dynamics and state-alignment losses. These predictors are conditioned on actions and serve a planner that searches over candidate actions at test time. Xu et~al.~\cite{xu2026jepapolicy} instead predict an action chunk jointly with a global embedding of a future frame in a shared transformer, with one 44M-parameter policy per task. Compared with these models, \method{} conditions its predictor on the task rather than on actions and computes actions without search. It supervises the forecast with dense patch features rather than a global embedding and shares one predictor across all tasks. Its target comes from a frozen encoder and therefore cannot collapse, which makes a regularizer such as SIGReg unnecessary.

\section{Method}\label{sec:method}

\method{} is a reactive policy that, at each query, forecasts the latent tokens of the scene given the task and decodes an action chunk from the current and forecast tokens (Fig.~\ref{fig:arch}).

\subsection{Problem Formulation}\label{sec:formulation}
We consider a closed set of manipulation tasks. At time $t$, the policy observes RGB images $o_t$ from several cameras, the robot state $s_t$ and the identity $k$ of the task, and outputs a chunk of $H$ actions $a_{t:t+H}$. \method{} factorizes it as
\begin{equation}
\begin{aligned}
\hat z_{t+h_1},\,\hat z_{t+h_2}&=F(z_t,\,g_k),\\
\hat a_{t:t+H}&=D(z_t,\,\hat z_{t+h_1},\,\hat z_{t+h_2},\,s_t,\,g_k),
\end{aligned}
\label{eq:factor}
\end{equation}
where $z_t$ are the visual tokens of the current images, $g_k$ is a code of task $k$, $F$ is the predictor, $D$ is the action decoder, and $h_1<h_2$ are two future horizons. As in JEPA world models~\cite{maes2026leworldmodel,liu2026groundedjepa}, $F$ predicts future representations rather than pixels. Unlike them, $F$ is conditioned on neither actions nor the robot state, and the policy therefore requires no search over candidate actions. The task code conditions both $F$ and $D$. Following LiLa\nobreakdash-WAM~\cite{yang2026lilawam}, $g_k$ is computed from the demonstrations of task $k$ and mapped by an MLP.

\subsection{Visual Encoding}\label{sec:enc}
Each camera view is encoded by a frozen DINOv3 ViT~\cite{simeoni2025dinov3}, whose dense self-supervised features preserve the spatial configuration of objects and the gripper~\cite{zhou2024dinowm,yang2026lilawam}. Following LiLa\nobreakdash-WAM, we fuse the patch features of three intermediate layers by layer normalization, concatenation and a linear projection. An adapter $A$ with 64 learned queries $q$ maps the fused features of each view to 64 tokens,
\begin{equation}
z=\overline{\mathrm{LN}}\big(A(q;U)\big)+e,
\label{eq:adapter}
\end{equation}
where $U$ are the fused features, $\overline{\mathrm{LN}}$ a layer normalization without affine parameters that normalizes the scale of the tokens, and $e$ a learned camera embedding. The tokens of all views are stacked into $z_t$ ($K{=}128$ for two cameras).

\subsection{Task-Conditioned Latent Predictor}\label{sec:foresight}
The predictor $F$ forecasts the tokens at both horizons in parallel rather than autoregressively. Its input consists of the $K$ current tokens and, for each horizon, $K$ query tokens initialized with a learned horizon vector. A temporal embedding, a positional embedding and the projected task code are added to each input token. A stack of bidirectional transformer blocks applies self-attention across all views and temporal positions. Each forecast is a residual update of the current tokens,
\begin{equation}
\hat z_{t+h_j}=\overline{\mathrm{LN}}\big(z_t+W_o\,r_j\big),\qquad j\in\{1,2\},
\label{eq:residual}
\end{equation}
where $r_j$ is the output of the last block at the queries of horizon $h_j$ and $W_o$ an output projection. With $W_o$ initialized to zero, $F$ initially forecasts a static scene and learns the change as a residual. The far horizon $h_2$ equals the chunk length $H$, and the near horizon $h_1{=}H/2$ provides the forecast scene at the midpoint of the chunk.

\subsection{Forecast Supervision}\label{sec:teacher}
We supervise the forecast in the patch-feature space of the frozen encoder, adopting the target of LiLa\nobreakdash-WAM. A patch decoder $G$, used only during training, maps the 64 forecast tokens of a view to $P$ patch features. The decoded features are compared with the patch features $X$ of the recorded future frame,
\begin{equation}
\mathcal{L}_{\mathrm{fut}}=\mathbb{E}_{j,v,p}\Big[1-\cos\big(G(\hat z_{t+h_j})_p,\,\mathrm{sg}(X_{t+h_j,p})\big)\Big],
\label{eq:lfut}
\end{equation}
where the expectation is taken over horizons $j$, views $v$ and patches $p$, $\mathrm{sg}(\cdot)$ denotes the stop-gradient operator, and frames beyond the end of an episode are masked. An analogous term $\mathcal{L}_{\mathrm{cur}}$ supervises the reconstruction of the current patch features from $z_t$. Two VICReg-style penalties~\cite{bardes2021vicreg}, $\mathcal{L}_{\mathrm{div}}$ and $\mathcal{L}_{\mathrm{var}}$, penalize the cosine similarity and the low variance among the tokens of each frame, for current and forecast tokens.

This target makes the forecast geometry-aware. A global embedding constrains only the overall content of the future frame, whereas location-specific patch features also constrain where the objects and the gripper will be. The forecast thus captures the spatial layout of each view without depth or 3D supervision (Fig.~\ref{fig:pca}). The target is computed by the frozen encoder rather than by the trainable adapter. It remains fixed during training and therefore cannot collapse, and $\mathcal{L}_{\mathrm{div}}$ and $\mathcal{L}_{\mathrm{var}}$ act only within each frame.

\subsection{Action Decoder and Training Objective}\label{sec:act}
The action decoder follows the ACT-style~\cite{zhao2023act} design of TurboVLA~\cite{xie2026turbovla}. A set of $H$ learned action queries $q^{a}$ attends to a memory $M_t$ that contains the current tokens, both forecasts, two state tokens and one task token. An MLP then regresses the entire chunk from the decoder outputs,
\begin{equation}
\begin{aligned}
M_t&=\big[z_t;\,\hat z_{t+h_1};\,\hat z_{t+h_2};\,\sigma(s_t);\,W_g\,g_k\big],\\
\hat a_{t:t+H}&=\tanh\big(\mathrm{MLP}(\mathrm{Dec}(q^{a};M_t))\big),
\end{aligned}
\label{eq:act}
\end{equation}
where $\mathrm{Dec}$ is a transformer decoder, $\sigma$ maps the state to two tokens and $W_g$ projects the task code to one token. The decoder is trained with an $\ell_1$ loss $\mathcal{L}_{\mathrm{act}}$ on the normalized actions. We adopt deterministic regression rather than a diffusion or flow-matching head. Regression requires no iterative sampling at inference and, for a compact multi-task policy on LIBERO, performs on par with flow matching~\cite{sendai2026minerva}.

All modules except the encoder are trained jointly from scratch by minimizing
\begin{equation}
\begin{aligned}
\mathcal{L}={}&\mathcal{L}_{\mathrm{act}}+\lambda_{\mathrm{fut}}\mathcal{L}_{\mathrm{fut}}+\lambda_{\mathrm{cur}}\mathcal{L}_{\mathrm{cur}}\\
&+\lambda_{\mathrm{reg}}(\mathcal{L}_{\mathrm{div}}+\mathcal{L}_{\mathrm{var}})+\lambda_{\mathrm{tmp}}\mathcal{L}_{\mathrm{tmp}},
\end{aligned}
\label{eq:total}
\end{equation}
where the $\lambda$ are loss weights and $\mathcal{L}_{\mathrm{tmp}}$, applied only on RoboTwin~2.0, penalizes the second temporal difference of the current and forecast tokens. The action loss also propagates through the forecast into the predictor and the adapter. The forecast is therefore shaped by both the future target and the control objective. At inference, the patch decoder is discarded, and each query processes only the current frame of every camera, without iterative sampling or search.

\begin{table*}[t]
\centering
\caption{Success Rates (\%) on LIBERO and RoboTwin~2.0}
\label{tab:main}
\scriptsize
\setlength{\tabcolsep}{3.5pt}
\begin{tabular}{l w{r}{34.8pt} *{9}{w{c}{34.8pt}}}
\toprule
& & & & \multicolumn{5}{c}{LIBERO} & \multicolumn{2}{c}{RoboTwin~2.0}\\
\cmidrule(lr){5-9}\cmidrule(l){10-11}
Method & Params & Robot PT & Task input & Spatial & Object & Goal & Long & Avg. & Clean & Rand.\\
\midrule
\grouprow{11}{Vision-language-action models}
OpenVLA~\cite{kim2024openvla} & 7.5B\tsup{*} & \cmark & Lang. & 84.7 & 88.4 & 79.2 & 53.7 & 76.5 & -- & --\\
$\pi_0$~\cite{black2025pi0} & 3.2B\tsup{*} & \cmark & Lang. & 96.8 & 98.8 & 95.8 & 85.2 & 94.2 & -- & --\\
$\pi_{0.5}$~\cite{intelligence2025pi05} & 3.4B & \cmark & Lang. & \second{98.8} & 98.2 & 98.0 & 92.4 & 96.9 & 82.74\tsup{a} & 76.76\tsup{a}\\
OpenVLA\nobreakdash-OFT~\cite{kim2025oft} & 7.6B & \cmark & Lang. & 97.7 & 98.0 & 96.1 & 95.3 & 96.8 & -- & --\\
VLA\nobreakdash-Adapter~\cite{wang2026vlaadapter} & 1.5B & \xmark & Lang. & 97.8 & 99.2 & 97.2 & 95.0 & 97.3 & -- & --\\
StarVLA Qwen3-OFT~\cite{community2026starvla} & 4.5B & \ns & Lang. & 97.8 & 98.6 & 96.2 & 93.8 & 96.6 & 88.18 & 88.32\\
VLAct-PI~\cite{yang2026vlact} & $\approx$4B\tsup{*} & \cmark & Lang. & -- & -- & -- & -- & -- & \second{93.0}\dz & 88.8\dz\\
\grouprow{11}{World-action models}
Fast\nobreakdash-WAM~\cite{yuan2026fastwam} & 11.9B & \xmark\tsup{b} & Lang. & 98.2 & \best{100.0} & 97.0 & 95.2 & 97.6 & 91.88 & \second{91.78}\\
LingBot\nobreakdash-VA~\cite{li2026lingbotva} & 10.9B & \cmark & Lang. & 98.5 & \second{99.6} & 97.2 & \best{98.5} & \second{98.5} & 92.93 & 91.55\\
LaWAM~\cite{chen2026lawam} & 2.1B & \cmark & Lang. & \best{99.4} & \second{99.6} & \best{98.4} & 97.0 & \best{98.6} & 92.64 & 89.80\\
Motus~\cite{bi2025motus} & 13.9B & \cmark & Lang. & -- & -- & -- & -- & -- & 88.66 & 87.02\\
\grouprow{11}{Compact policies ($<$1B)}
Octo-Base~\cite{ghosh2024octo} & 93M\tsup{*} & \cmark & Lang. & 78.9 & 85.7 & 84.6 & 51.1 & 75.1 & -- & --\\
SmolVLA~\cite{shukor2025smolvla} & 0.40B & \xmark & Lang. & 90.0 & 96.0 & 92.0 & 71.0 & 87.3\tsup{c} & -- & --\\
Evo\nobreakdash-1~\cite{lin2025evo1} & 0.78B & \xmark & Lang. & 92.7 & 97.7 & 96.3 & 92.3 & 94.8\tsup{c} & -- & --\\
X\nobreakdash-VLA~\cite{zheng2025xvla} & 0.88B & \cmark & Lang. & 98.2 & 98.6 & 97.8 & \second{97.6} & 98.1 & 72.80\tsup{a} & 72.84\tsup{a}\\
TurboVLA\smash{$^{d}$}~\cite{xie2026turbovla} & 0.2\tsup{*}/0.43B & \xmark & Lang. & 97.0 & \best{100.0} & \second{98.2} & 95.2 & 97.6 & 88.02 & 88.10\\
LiLa\nobreakdash-WAM\smash{$^{e}$}~\cite{yang2026lilawam} & 0.48B & \xmark & ID & 98.0 & 98.8 & 97.2 & 94.2 & 97.1 & 90.48\tsup{c} & 89.04\tsup{c}\\
\midrule
\ours \method{}-S\smash{$^{\dagger f}$} (ours) & 38/77M & \xmark & ID & 96.2 & 99.8 & 95.0 & 91.2 & 95.6 & \best{93.94} & \best{93.10}\\
\bottomrule
\end{tabular}\\[2pt]
\parbox{\textwidth}{\scriptsize
Params: parameters of the modules that compute an action chunk from the observation and the instruction, including frozen image encoders, text encoders and VAE encoders; modules used only in training or only for outputs that the actions do not depend on (e.g., VAE decoders, language-model heads) are excluded. Counted from the released configurations, except values marked with $^{*}$. Robot PT: pretraining on robot data or embodied videos. Task input: the task is specified by a language instruction (Lang.) or by a task identifier (ID). RoboTwin~2.0 results follow the mixed protocol, in which one multi-task policy is trained on 50 clean and 500 randomized demonstrations per task. The LIBERO results of Octo are taken from OpenVLA~\cite{kim2024openvla}, those of $\pi_0$ from OpenVLA\nobreakdash-OFT~\cite{kim2025oft} and those of $\pi_{0.5}$ from its public repository. Bold: best; underlined: second best (LIBERO columns exclude the $\dagger$ row). --: not reported; \ns: not stated.\\
$^{*}$Reported by the authors or by TurboVLA~\cite{xie2026turbovla}; not recounted. $^{a}$Reproduced by LingBot\nobreakdash-VLA~\cite{wu2026lingbotvla} ($\pi_{0.5}$) and by Motus~\cite{bi2025motus} (X\nobreakdash-VLA). $^{b}$Initialized from a pretrained video generator. $^{c}$Fewer evaluation episodes: 10 per task for SmolVLA and Evo\nobreakdash-1, and 50 per task for LiLa\nobreakdash-WAM on RoboTwin~2.0. $^{d}$0.2B on LIBERO and 0.43B on RoboTwin~2.0, where it uses three cameras and is evaluated with unseen instructions. $^{e}$One policy per LIBERO suite; one third-person camera on RoboTwin~2.0. $^{f}$38M on LIBERO and 77M on RoboTwin~2.0 (two configurations, Sec.~\ref{sec:setup}). $^{\dagger}$Trained on the Extended LIBERO data (Sec.~\ref{sec:setup}).}
\end{table*}

\section{Experiments}\label{sec:experiments}

Section~\ref{sec:setup} describes the setup. We then compare \method{} with published VLAs, WAMs and compact policies in success rate, model size and computational cost (Sec.~\ref{sec:main}). In Sec.~\ref{sec:ablation}, an ablation quantifies the contribution of the world model, and two analyses show what the forecast encodes and how the action decoder attends to it. Section~\ref{sec:realrobot} evaluates \method{} on a real robot.

\subsection{Experimental Setup}\label{sec:setup}

\textbf{Benchmarks.} LIBERO~\cite{liu2023libero} consists of four task suites (Spatial, Object, Goal and Long), each containing ten tasks. A single policy is trained on all 40 tasks. The \emph{Standard} setting consists of the 1{,}693 no-op-filtered demonstrations released with OpenVLA~\cite{kim2024openvla}, as in prior work. The \emph{Extended} setting adds in-domain data, for a total of 6{,}585 trajectories (1.03M frames): the 90 tasks of LIBERO\nobreakdash-90, 862 official demonstrations discarded by the OpenVLA filter and re-rendered from their recorded simulator states, 21 replayed LIBERO\nobreakdash-Long demonstrations and 89 successful rollouts of a StarVLA Qwen3\nobreakdash-VL\nobreakdash-OFT policy~\cite{community2026starvla}. Results with the Extended data are marked with $\dagger$ in Table~\ref{tab:main}. We evaluate all 50 initial states of each task (2{,}000 episodes).

RoboTwin~2.0~\cite{chen2025robotwin2} comprises 50 bimanual tasks on the Aloha-AgileX platform, and one policy is again trained on all tasks. In the \emph{mixed} protocol of recent VLAs and WAMs~\cite{bi2025motus,xie2026turbovla,yuan2026fastwam}, the policy is trained on 50 clean and 500 domain-randomized demonstrations per task and evaluated in clean scenes and in randomized scenes with unseen textures, clutter, lighting and table heights. We adopt the data release, outlier list and action space of LiLa\nobreakdash-WAM~\cite{yang2026lilawam}: 27{,}071 demonstrations, head and front cameras at 320$\times$240 and 14-D absolute joint targets. Each setting is evaluated with 100 episodes per task (5{,}000 in total). Our simulator build fixes the renderer's denoiser and inverse-kinematics checks without changing the task logic.

\textbf{Baselines.} We compare \method{} with published results of VLAs, WAMs and compact policies, with sources given in Table~\ref{tab:main}. \method{}\nobreakdash-S, the default model, builds on a frozen ViT\nobreakdash-S/16 encoder and is evaluated at its final checkpoint.

\textbf{Ablation protocol.} We refer to the predictor together with all feature losses as the world model. The variant without the world model retains the encoder, the adapter, the action decoder and the task code, but omits the forecast from the decoder memory and all feature losses ($\mathcal{L}_{\mathrm{fut}}$, $\mathcal{L}_{\mathrm{cur}}$, $\mathcal{L}_{\mathrm{div}}$, $\mathcal{L}_{\mathrm{var}}$, $\mathcal{L}_{\mathrm{tmp}}$). It reduces to an ACT-style policy trained with $\mathcal{L}_{\mathrm{act}}$ on the same frozen features. We compare both models in three regimes: LIBERO with a quarter of the Standard demonstrations (427, 7\nobreakdash--13 per task), LIBERO with the Extended data, and RoboTwin~2.0 under the mixed protocol with a shortened schedule of 191k instead of 765k training steps.

\textbf{Implementation.} \method{}\nobreakdash-S fuses layers 7, 9 and 11 of the 12-layer DINOv3 ViT\nobreakdash-S/16 (21.6M parameters). The LIBERO configuration has a two-layer adapter of width 384, a two-block predictor and a two-layer action decoder. Its inputs are 256$\times$256 images, and an action chunk has $H{=}8$ steps of 6-D end-effector deltas and a gripper command, with ${(h_1,h_2)=(4,8)}$. The RoboTwin~2.0 configuration has a four-layer adapter of width 768, a four-block predictor and a three-layer action decoder, with a 16-D end-effector state, $H{=}32$ and ${(h_1,h_2)=(16,32)}$; the first 16 steps are executed after cubic B\nobreakdash-spline smoothing. We set $\lambda_{\mathrm{fut}}{=}1$, $\lambda_{\mathrm{cur}}{=}0.1$, $\lambda_{\mathrm{reg}}{=}0.02$, and $\lambda_{\mathrm{tmp}}{=}0$ on LIBERO and $0.1$ on RoboTwin~2.0. Following LiLa\nobreakdash-WAM, we train with AdamW at a learning rate of $2{\times}10^{-4}$ and a two-phase cosine schedule: 80k steps at batch size 160 on 8 GPUs for LIBERO and 765k steps at batch size 128 on 32 GPUs for RoboTwin~2.0 (NVIDIA RTX PRO 5000). On RoboTwin~2.0, the decoder receives the task code in an additional fine-tuning stage of 9.6k steps that starts from the 765k-step model, with the learning rate decaying from $4{\times}10^{-5}$ to $10^{-5}$. This stage changes success by $-$0.4 points in both settings, with intervals that include zero. The code builds on StarVLA~\cite{community2026starvla}.

\textbf{Statistics.} We report success rates with Wilson 95\% confidence intervals and, for differences between models, paired per-task bootstrap intervals.

\begin{table}[t]
\centering
\caption{Ablation of the World Model}
\label{tab:ablation}
\scriptsize
\setlength{\tabcolsep}{2.0pt}
\begin{tabular}{@{}l c c c c c c c@{}}
\toprule
& \multicolumn{5}{c}{LIBERO, $\frac{1}{4}$ data} & LIBERO & RoboTwin\\
\cmidrule(lr){2-6}
& Spatial & Object & Goal & Long & Avg. & Extended & 191k steps\\
\midrule
\ours \method{}-S & \best{90.2} & 85.4 & \best{87.8} & \best{58.4} & \best{80.5} & \best{95.6} & \best{90.7}\\
w/o world model & 77.2 & \best{89.0} & 80.2 & 37.4 & 71.0 & 93.1 & 18.8\\
\midrule
$\Delta$ & $+$13.0 & $-$3.6 & $+$7.6 & $+$21.0 & $+$9.5 & $+$2.5 & $+$71.9\\
95\% CI & & & & & [5.2, 14.2] & [$-$0.8, 6.1] & [65.2, 78.0]\\
\bottomrule
\end{tabular}\\[2pt]
\parbox{\columnwidth}{\scriptsize Success rates (\%) of \method{}\nobreakdash-S with and without the world model. On LIBERO, the models are trained on a quarter of the Standard demonstrations or on the Extended data and evaluated on 2{,}000 episodes. On RoboTwin~2.0, they are trained under the mixed protocol for 191k instead of 765k steps and evaluated on 100 clean episodes per task. $\Delta$: paired per-task difference; 95\% CI: bootstrap confidence interval of $\Delta$.}
\end{table}

\subsection{Main Results}\label{sec:main}
\textbf{RoboTwin~2.0.} \method{} outperforms billion-parameter WAMs on this benchmark. \method{}\nobreakdash-S achieves 93.94\% in clean and 93.10\% in randomized scenes with 77M parameters (Table~\ref{tab:main}; 95\% intervals [93.2,~94.6] and [92.4,~93.8]). The four strongest baselines, LaWAM (2.1B), VLAct-PI (${\approx}$4B), LingBot\nobreakdash-VA (10.9B) and Fast\nobreakdash-WAM (11.9B), attain 91.9\nobreakdash--93.0\% in clean and 88.8\nobreakdash--91.8\% in randomized scenes, all below the lower bounds of these intervals. Among compact policies, \method{}\nobreakdash-S exceeds TurboVLA (0.43B) by 5.9 and 5.0 points and LiLa\nobreakdash-WAM (0.48B) by 3.5 and 4.1 points.

\textbf{LIBERO.} \method{}\nobreakdash-S attains 95.6\% with 38M parameters, above $\pi_0$ (3.2B, 94.2\%) and Evo\nobreakdash-1 (0.78B, 94.8\%). It is 2.5 points below X\nobreakdash-VLA (0.88B), the strongest compact policy. Section~\ref{sec:ablation} quantifies the contribution of the world model.

\textbf{Model size and cost.} \method{}\nobreakdash-S is the smallest model in Table~\ref{tab:main} and Fig.~\ref{fig:teaser}, with 77M parameters on RoboTwin~2.0 and 38M on LIBERO. The other compact policies have 93M\nobreakdash--0.88B parameters, and the VLAs and WAMs 1.5\nobreakdash--13.9B. \method{} uses neither a VLM nor a video generator. The frozen ViT\nobreakdash-S/16 encoder accounts for 21.6M parameters and the predictor of the world model for 4.0M on LIBERO and 6.0M on RoboTwin~2.0. The 5.1M-parameter patch decoder is used only in training and is not counted.

A larger encoder does not improve success. With ViT\nobreakdash-L, success on RoboTwin~2.0 is 93.78\% and 93.42\% in clean and randomized scenes, compared with 94.36\% and 93.48\% for the 765k-step \method{}\nobreakdash-S model. On the Extended LIBERO data, ViT\nobreakdash-S differs from ViT\nobreakdash-L by $-$0.5 points [$-$3.5,~2.5].

Training \method{}\nobreakdash-S requires 45 GPU-hours on LIBERO and 1{,}749 on RoboTwin~2.0, where ViT\nobreakdash-L requires 2{,}903.

\begin{figure}[t]
\centering
\includegraphics[width=\columnwidth]{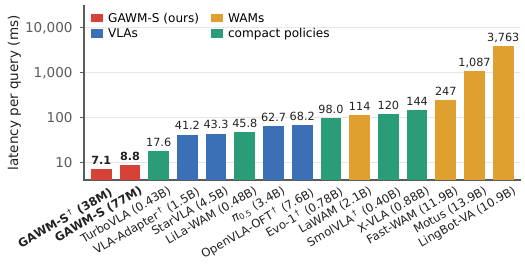}
\caption{\textbf{Inference latency per action-chunk query.} Official implementations on the same GPU model (NVIDIA RTX PRO 5000) with batch size 1 and default precision and denoising steps. Latency covers the modules counted in Table~\ref{tab:main}, from preprocessed inputs to the action chunk (median of 200 queries). $\dagger$: LIBERO configuration; all others use RoboTwin~2.0. Parentheses: parameters as in Table~\ref{tab:main}. Baselines are randomly initialized, which does not change the computation.}
\label{fig:latency}
\end{figure}

\textbf{Inference latency.} \method{}\nobreakdash-S has the lowest latency of all compared methods (Fig.~\ref{fig:latency}). One action chunk takes 8.8~ms on RoboTwin~2.0 and 7.1~ms on LIBERO, compared with 17.6\nobreakdash--143.5~ms for the other compact policies, 41.2\nobreakdash--68.2~ms for the VLAs and 114\nobreakdash--3{,}763~ms for the WAMs. \method{}\nobreakdash-S predicts the chunk in a single forward pass, whereas most baselines sample actions over 10 or more denoising steps and video-based WAMs also run a video backbone per query.

\subsection{Ablation and Analysis}\label{sec:ablation}
\textbf{World-model ablation.} The world model substantially improves success when training data or training time is limited (Table~\ref{tab:ablation}). With a quarter of the LIBERO demonstrations, removing it lowers success by 9.5 points [5.2,~14.2]. The drop is largest on LIBERO\nobreakdash-Long (21.0 points), whose tasks comprise several subtasks. On LIBERO\nobreakdash-Object, whose tasks differ mainly in the object to be grasped, the variant without the world model is 3.6 points higher. On RoboTwin~2.0 with the shortened 191k-step schedule, removing the world model reduces success from 90.7\% to 18.8\%, and \method{}\nobreakdash-S outperforms the variant on all 50 tasks. This gap suggests that, within this schedule, the action loss alone does not shape the adapter tokens into a representation that supports bimanual control. With the Extended LIBERO data, the gain is 2.5 points [$-$0.8,~6.1]. The gain from the world model thus increases as training data become scarce.

\begin{figure*}[t]
\centering
\includegraphics[width=\textwidth]{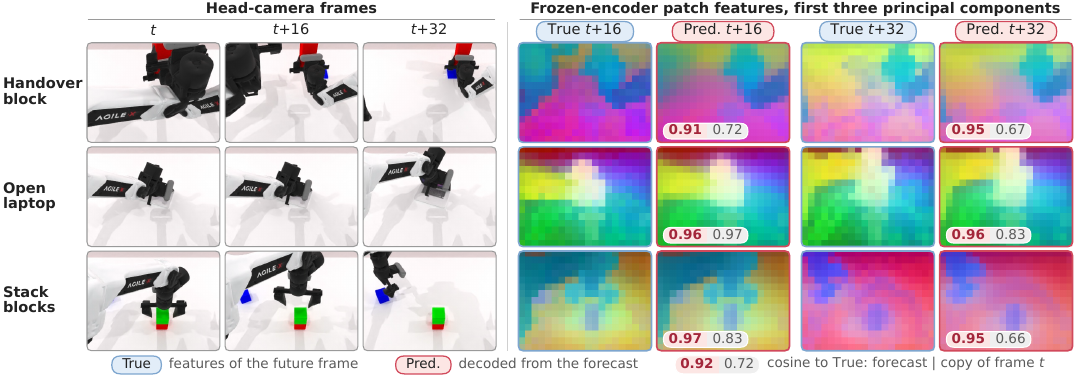}
\caption{\textbf{Decoded forecasts.} \method{}\nobreakdash-S on RoboTwin~2.0, head camera. Left: frames $t$, $t{+}16$ and $t{+}32$. Right: first three principal components of the encoder patch features of the true future frames (True) and of the decoded forecast (Pred.), projected on a PCA basis shared within each example. Labels: mean per-patch cosine similarity to the true features for the forecast and for frame $t$. Examples: first clean demonstrations, at the frame with the largest 32-frame feature change within 20\nobreakdash--80\% of the episode.}
\label{fig:pca}
\end{figure*}

\textbf{Forecast quality.} The decoded forecast captures the spatial layout of the future scene more accurately than the current frame does (Fig.~\ref{fig:pca}). We sample 1{,}000 anchor frames from the RoboTwin~2.0 demonstrations of all 50 tasks and both cameras. The mean per-patch cosine similarity between the decoded forecast and the true future features is 0.970 at $+$16 frames and 0.967 at $+$32 frames, compared with 0.929 and 0.892 for the current frame. The forecast is closer to the true future than the current frame on 75\% and 91\% of the anchors, respectively. Its advantage grows with the horizon as the scene departs further from the current frame. The forecast is smoother than the target. Its similarity approaches 0.983, the bound of reconstructing the current frame from 64 tokens.

\begin{figure}[t]
\centering
\includegraphics[width=\columnwidth]{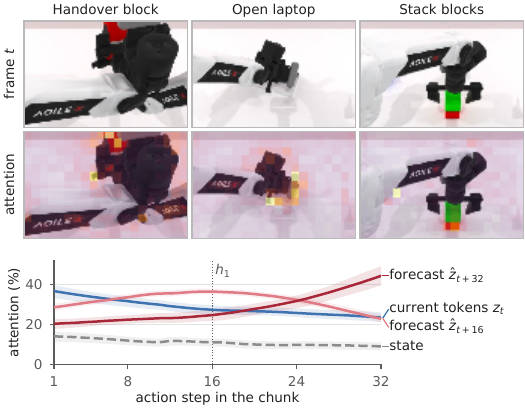}
\caption{\textbf{Attention of \method{}\nobreakdash-S on RoboTwin~2.0.} Top: head-camera frame $t$ of the examples of Fig.~\ref{fig:pca} and the adapter cross-attention over its patches (mean over queries, heads and layers; brighter is higher). Bottom: share of the action decoder's cross-attention on each memory group (current tokens, near and far forecasts, state tokens) per action step (mean over heads, layers and 1{,}000 anchors; bands: interquartile range).}
\label{fig:attention}
\end{figure}

\textbf{Attention.} Each action step attends most to the tokens closest to it in time (Fig.~\ref{fig:attention}, bottom). Over the same 1{,}000 anchors, the first steps of the chunk attend mostly to the current tokens, the steps around $h_1$ to the near forecast and the last steps to the far forecast. The share of the far forecast rises from 20\% at the first step to 44\% at the last. This progression holds for every anchor. The adapter attention is spread over the scene, with its peaks on the gripper fingers and the manipulated objects (Fig.~\ref{fig:attention}, top).

\subsection{Real-Robot Evaluation}\label{sec:realrobot}

\begin{figure}[t]
\centering
\includegraphics[width=\columnwidth]{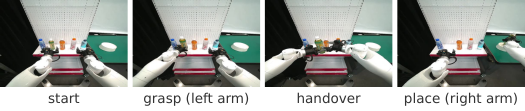}
\caption{\textbf{Real-robot task.} Head-camera frames of a demonstration of the green-bottle task. The four tasks differ only in the target bottle.}
\label{fig:robot}
\end{figure}

\textbf{Platform and data.} The platform is a Tianji Marvin Pro robot with two 7-DoF arms, each equipped with a parallel gripper, as well as a head camera and two wrist cameras. In each of four tasks, the left arm grasps a target bottle (blue, yellow, green or black) from a shelf holding five bottles and hands it to the right arm, which places it on the table (Fig.~\ref{fig:robot}). The scene is identical across tasks, and the task code is therefore the only cue to the target bottle. We collected 340 teleoperated demonstrations at 15~Hz (113k frames; 74\nobreakdash--93 per task): 307 for training and 33 for checkpoint selection. A further 81 demonstrations (21k frames), recorded on a separate day, form the test set. \method{}\nobreakdash-S adopts the RoboTwin~2.0 configuration with all three cameras and predicts chunks of 16 absolute joint and gripper targets. It is initialized from the 765k-step RoboTwin~2.0 model and fine-tuned for 40k steps. The baseline is $\pi_{0.5}$~\cite{intelligence2025pi05}, fine-tuned with LoRA on all 340 demonstrations with language instructions and 20-step chunks: 40k steps at $5{\times}10^{-6}$ and 60k at $2.5{\times}10^{-5}$.

\begin{table}[t]
\centering
\caption{Evaluation on the Real Robot}
\label{tab:realrobot}
\scriptsize
\setlength{\tabcolsep}{3.5pt}
\begin{tabular}{@{}l >{\columncolor{oursbg}}c c c@{}}
\toprule
& \method{}-S & $\pi_{0.5}$ (40k) & $\pi_{0.5}$ (60k)\\
\midrule
Joint error, first step & \best{0.072} & 0.098 & 0.092\\
Joint error, full chunk & \best{0.088}\tsup{a} & 0.106\tsup{b} & 0.101\tsup{b}\\
Gripper error, first step & \best{0.020} & 0.028 & 0.023\\
Closed-loop success, yellow bottle & \best{46/50}\tsup{c} & -- & 39/50\\
Latency, median / 95th pct.\ (ms) & \best{9.3} / \best{9.6} & \multicolumn{2}{c}{63.8 / 63.9}\\
\bottomrule
\end{tabular}\\[2pt]
\parbox{\columnwidth}{\scriptsize Errors: mean absolute errors on the 81 test demonstrations with respect to the recorded joint targets of both arms (rad) and gripper commands. Success: closed-loop trials in which the bottle is placed on the table. Latency: batch size 1 on the same RTX PRO 5000 GPU, measured as in Fig.~\ref{fig:latency}.\\
$^{a}$Over 16 steps. $^{b}$Over 20 steps. $^{c}$Fine-tuned for 60k steps.}
\end{table}

\textbf{Results.} Offline, we compare the predicted actions with the recorded ones on each test demonstration. \method{}\nobreakdash-S predicts the first action of each chunk with a joint error of 0.072~rad (4.1$^\circ$), 26\% and 22\% lower than the two $\pi_{0.5}$ models, and achieves the lowest gripper error (Table~\ref{tab:realrobot}). Over the full chunk, its error (0.088~rad, 16 steps) is also lower than that of $\pi_{0.5}$ (0.101\nobreakdash--0.106~rad, 20 steps). In 50 closed-loop trials on the yellow-bottle task, \method{}\nobreakdash-S succeeds 46 times (92\%, [81.2, 96.8]) and $\pi_{0.5}$ (60k) 39 times (78\%, [64.8, 87.2]). With 77M rather than 3.4B parameters, \method{}\nobreakdash-S requires 9.3~ms per query and $\pi_{0.5}$ 63.8~ms.

\section{Conclusion}\label{sec:conclusion}
We presented \method{}, a compact world-action model for reactive multi-task manipulation. A lightweight predictor forecasts the latent tokens of a frozen encoder at two future horizons. An action decoder then regresses action chunks directly from the current and forecast tokens. Without a vision-language model or a video generator, \method{} outperforms billion-parameter WAMs on RoboTwin~2.0 and performs comparably to billion-parameter VLAs such as $\pi_0$ on LIBERO. On a dual-arm robot, it achieves lower action-prediction error on held-out demonstrations than a fine-tuned billion-parameter VLA, with substantially lower inference latency. These results suggest that, on closed task sets, a compact task-conditioned latent predictor can provide a reactive policy with the foresight that existing WAMs obtain from billion-parameter backbones. This foresight encodes the future layout of the scene and substantially improves success when training data or training time is limited.

\textbf{Limitations and future work.} \method{} targets a closed set of tasks, each identified by a code derived from its demonstrations, and does not interpret language instructions. We plan to combine \method{} with a large language model that maps open-ended instructions to the known tasks, as in hierarchical robot systems~\cite{ahn2022saycan,shi2025hirobot}, adding language-level flexibility while retaining the efficiency of the compact policy.

\microtypesetup{protrusion=false}
\bibliographystyle{IEEEtran}
\bibliography{ieeetran_ctl,refs}

\end{document}